\pdfoutput=1

\documentclass[11pt]{article}
\usepackage[]{latex/style}
\usepackage{times}
\usepackage{latexsym}
\usepackage[T1]{fontenc}
\usepackage[utf8]{inputenc}
\usepackage{microtype}
\usepackage{inconsolata}
\usepackage{graphicx}
\usepackage{amssymb}
\usepackage{amsmath}
\usepackage{enumitem}
\usepackage{booktabs}
\usepackage{multirow}
\usepackage[normalem]{ulem}
\useunder{\uline}{\ul}{}

\title{Multimodal Prompt Learning with Missing Modalities for Sentiment Analysis and Emotion Recognition}

\author{Zirun Guo$^{1,2}$, Tao Jin$^{1}$\thanks{\quad Corresponding author} , Zhou Zhao$^{1,2}$ \\
$^1$ Zhejiang University, $^2$ Shanghai Artificial Intelligence Laboratory \\
\texttt{\{gzr,jint\_zju,zhaozhou\}@zju.edu.cn}}

\begin{document}
\maketitle
\begin{abstract}
The development of multimodal models has significantly advanced multimodal sentiment analysis and emotion recognition. However, in real-world applications, the presence of various missing modality cases often leads to a degradation in the model's performance. In this work, we propose a novel multimodal Transformer framework using prompt learning to address the issue of missing modalities. Our method introduces three types of prompts: generative prompts, missing-signal prompts, and missing-type prompts. These prompts enable the generation of missing modality features and facilitate the learning of intra- and inter-modality information. Through prompt learning, we achieve a substantial reduction in the number of trainable parameters. Our proposed method outperforms other methods significantly across all evaluation metrics. 
Extensive experiments and ablation studies are conducted to demonstrate the effectiveness and robustness of our method, showcasing its ability to effectively handle missing modalities. Codes are available at \url{https://github.com/zrguo/MPLMM}.
\end{abstract}

\section{Introduction}

Humans perceive the world in a multimodal way, such as sight, sound, touch and language.
These multimodal features can provide comprehensive information to help us understand and explore the world. Thus, modeling and mining multimodal data is of great importance and has much potential. Recently, multimodal sentiment analysis~\citep{tsai-etal-2019-multimodal, hazarika2020misa, han-etal-2021-improving, hu-etal-2022-unimse} has attracted much attention. However, there are two main challenges in many existing methods: 
1) Different from common multimodal tasks which only have two modalities (image and text), multimodal sentiment analysis task often has more modalities (video, audio, text, etc.). Therefore, in real-world scenarios, missing modality conditions always occur due to equipment failure, data corruption, privacy issues and the like, especially in low-resource domains, which could lead to a degradation in the model's performance. Current multimodal models trained on complete data usually fail when tested on incomplete data~\citep{aguilar-etal-2019-multimodal, pham2020translation}.
2) With the success of large-scale multimodal models~\citep{kim2021vilt, ALBEF, radford2021learning}, lots of researchers tend to finetune these large pre-trained models to downstream tasks. However, this kind of finetuning is infeasible for many researchers because it requires large computational resources. Besides, finetuning such a pre-trained model on small datasets could lead to instability~\citep{mosbach2021on}.

Recently, prompt learning~\citep{gao-etal-2021-making, heinzerling-inui-2021-language, khattak2023maple, lee2023multimodal} is proposed, which freezes all the parameters of a pre-trained model while only finetuning several prompts and it has achieved great success~\citep{lester-etal-2021-power}. Motivated by prompt learning, in this paper, we intend to exploit a high-resource dataset that contains relatively more complete modality data for pre-training and then leverage several trainable prompts to transfer the knowledge from high-resource domains to low-resource domains where missing modality cases often occur.

Previous works~\citep{ma2021smil, pham2020translation, zhao-etal-2021-missing} mainly focus on introducing sophisticated architecture to address the issue of missing modalities. These methods do not use pre-trained models and usually require a lot of computational resources. However, our method is based on prompt learning, which only finetunes a few parameters of prompts. \citet{lee2023multimodal} is a recent work which is similar to ours. However, its proposed missing-aware prompts increase exponentially with the number of modalities. In contrast, our proposed prompts increase linearly with the number of modalities which is more parameter-efficient. Specifically, we propose three types of prompts: generative prompts, missing-signal prompts, and missing-type prompts which can learn the representations of the missing modalities, cross-modal and fine-grained features. These three types of prompts play a combined role in improving the model's performance.

We conduct extensive experiments on four datasets: CMU-MOSEI~\citep{bagher-zadeh-etal-2018-multimodal}, CMU-MOSI~\citep{7742221}, IEMOCAP~\citep{Busso2008IEMOCAPIE} and CH-SIMS~\citep{yu-etal-2020-ch}. The proposed method outperforms the baselines significantly across all metrics on all datasets. We further study the roles of three types of prompts, the effect of missing rate of training data, and the effect of prompt length. We find that: 
1) missing-signal prompts are modality-specific while missing-type prompts are modality-shared which represent intra-modality and inter-modality information respectively.
2) with short prompts, our model can achieve very good results which demonstrates our proposed method is parameter-efficient.
3) the missing rate is important for the performance of the model, with 70\% being the optimal value.

Our contributions can be summarized as follows:
\begin{itemize}[itemsep=0pt,parsep=0pt,topsep=0pt,partopsep=0pt]
  \item We present a novel framework via prompt learning for sentiment analysis and emotion recognition which is not only computationally efficient but also capable of handling missing modalities during both the training and testing stages.
  \item The number of parameters of our proposed prompts is linearly related to the number of modalities, which significantly reduces computational resources.
  \item We propose three types of prompts to address the issue of missing modalities. These three types of prompts can generate missing information, and learn intra- and inter-modality information respectively.
  \item Our proposed method outperforms all the baselines across all metrics significantly. Furthermore, we discover that applying modality dropout with a rate of 70\% during training yields the best enhancement in the model's performance.
\end{itemize}

\section{Related Works}
\textbf{Multimodal Sentiment Analysis (MSA) and Emotion Recognition (MER).} Multimodal sentiment analysis and emotion recognition refer to the process of analyzing and understanding human sentiment or emotions using multiple modalities of data, such as text, image, audio, and video. The main challenge of such tasks is how to effectively use the information from different modalities to complement each other. Currently, there are two main multimodal fusion strategies: feature-level fusion and decision-level fusion. Feature-level fusion methods~\citep{liang-etal-2018-multimodal, wang2018words} combine features from different modalities to create a unified feature representation via concatenation or other methods. For example, \citet{liang-etal-2018-multimodal} decomposed the fusion problem into multiple stages and fused features step by step to obtain a comprehensive representation. \citet{mai-etal-2019-divide} conducted fusion hierarchically so that both local and global interactions are considered for a comprehensive interpretation of multimodal embeddings. Different from feature-level fusion methods, decision-level fusion methods~\citep{tsai-etal-2019-multimodal, hazarika2020misa, han-etal-2021-improving, hu-etal-2022-unimse} process different modalities independently and then incorporate them into the final decision. For instance, \citet{tsai-etal-2019-multimodal} proposed a directional pairwise cross-modal attention to implement modal alignment and fused the outputs of each modality at the decision level to make predictions. These methods all assume that the data is complete while our proposed method can deal with the situation when there exist missing modalities.

\begin{figure*}
  \centering
  \includegraphics[width=\linewidth]{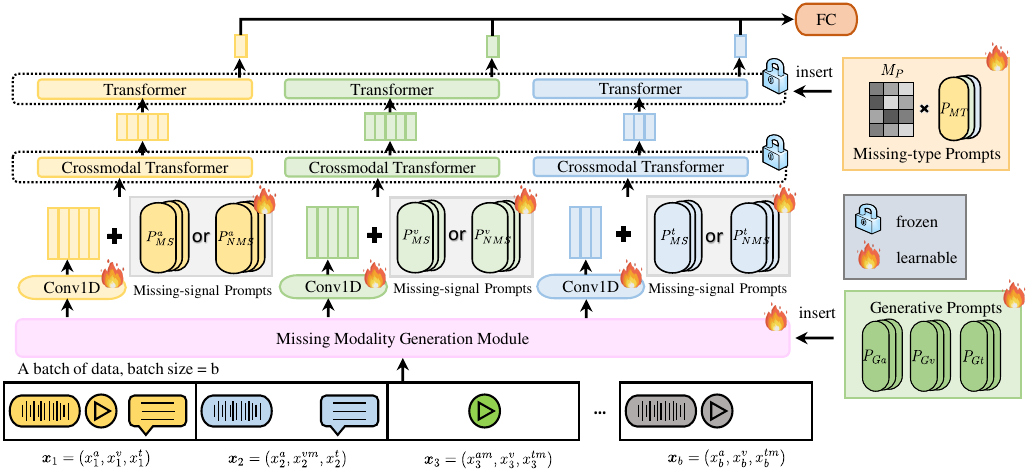}
  \caption{The overall architecture of our proposed method. A batch of data that contains different missing modality cases is fed to the Missing Modality Generation Module (see Section~\ref{s32}) to obtain generated features. They are then passed to the pre-trained backbone with missing-signal prompts and missing-type prompts (see Section~\ref{s33}).}
  \label{fig1}
\end{figure*}

\noindent\textbf{Multimodal Learning with Missing Modalities.} The presence of a missing modality poses challenges for multimodal learning because the model needs to effectively handle the absence of information while still making accurate predictions. \citet{ma2021smil} proposed the SMIL model which leverages Bayesian meta-learning to address the issue of missing modalities. Some methods~\citep{10.1145/3219819.3219963, Du_2018} directly generate missing modalities using the available modalities. \citet{zhao-etal-2021-missing} proposed learning robust joint multimodal representations which can predict the representation of any missing modality given the available modalities. However, these methods always introduced sophisticated architecture to address the issue of missing modalities, which is computationally expensive. In comparison, our approach utilizes three different prompts to handle missing modalities, which is computationally more efficient. In a more recent work~\citep{lee2023multimodal}, prompts are used to address missing modalities, but the number of prompts increases exponentially with the number of modalities. In contrast, the number of prompts in our method is linearly related to the number of modalities.

\noindent\textbf{Prompt Learning.} Prompt learning, which refers to the process of designing or generating effective prompts to use a pre-trained model for different types of downstream tasks, has been widely used in various NLP tasks~\citep{gao-etal-2021-making, heinzerling-inui-2021-language}. With the success of prompt learning in NLP tasks~\citep{lester-etal-2021-power, li-liang-2021-prefix, liu-etal-2022-p}, recent works~\citep{tsimpoukelli2021multimodal, liang-etal-2022-modular, khattak2023maple} explored to leverage prompts in multimodal learning. \citet{tsimpoukelli2021multimodal} presented a method for transforming large language models into multimodal systems by extending the soft-prompting philosophy of prefix tuning to ordered sets of images and texts. \citet{khattak2023maple} proposed a strategy to ensure synergy between vision-language modalities by explicitly conditioning the vision prompts on textual prompts across different Transformer stages. More recently, \citet{lee2023multimodal} proposed missing-aware prompts to address missing modalities which increase the robustness of the model, but it did not recover the missing information from the multimodal input. In comparison, our approach utilizes generative prompts to generate the representation of missing modalities given available modalities which can help further boost the performance of the model.

\section{Proposed Method}
In this section, we describe our proposed method (Figure~\ref{fig1}) via prompt learning to address the issue of missing modalities (introduced in Section~\ref{s31}). Specifically, we introduce three kinds of prompts: generative prompts (introduced in Section~\ref{s32}), missing-signal prompts, and missing-type prompts (introduced in Section~\ref{s33}).

\subsection{Overall Architecture}\label{s31}
\textbf{Problem Definition.}\quad Given a multimodal dataset $\mathcal{D}$ consisting of $M=3$ modalities (e.g., audio, video and text), we use $\boldsymbol{x}=(x^a,x^v,x^t)$ to represent a pair of features in $\mathcal{D}$, where $x^a,x^v,x^t$ represent the features of acoustic, visual and textual modalities respectively. To indicate missing modalities, we use $x^{am},x^{vm},x^{tm}$ to denote which modalities are absent.

Figure~\ref{fig1} shows the overall architecture of our proposed model. For simplicity and better comparison, we use MulT~\citep{tsai-etal-2019-multimodal} as the backbone, which introduced the Crossmodal Transformer for modeling unaligned data. In our proposed method, we employ three types of different prompts: generative prompts, missing-signal prompts, and missing-type prompts. The generative prompts assist the available modalities in generating representations for the missing modalities. The missing-signal prompts are designed to inform the model about the absence of a specific modality while the missing-type prompts inform the model about the absence of other modalities.

\begin{figure}
  \centering
  \includegraphics[width=\linewidth]{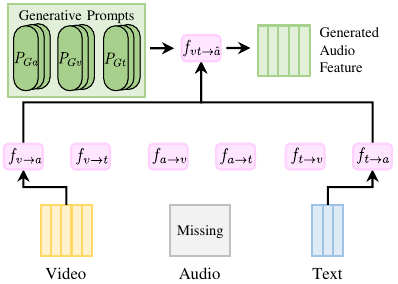}
  \caption{The illustration of Missing Modality Generation Module (MMGM). The figure shows the process of generating the audio feature of an example of $\boldsymbol{x}=(x^{am},x^v,x^t)$ where the audio modality is missing and the other two are not. It can be described using the Equation~\ref{e1}.}
  \label{fig2}
\end{figure}

\subsection{Missing Modality Generation Module (MMGM)}\label{s32}
Many methods address missing modality issues by recovering missing information using available modalities~\citep{10.1145/3219819.3219963, Du_2018}. However, these methods often utilize complex structures. Based on this observation, we propose the Missing Modality Generation Module (MMGM) which utilizes generative prompts to recover missing information in a much simpler way. We denote generative prompts as $\boldsymbol{P_G}=(P_{Ga}, P_{Gv}, P_{Gt})$ where $P_{Ga}, P_{Gv}$ and $P_{Gt}$ represent the generative prompts for the audio, video and text modalities, respectively. $\boldsymbol{P_G}\in \mathbb{R}^{3\times d_p\times \ell_p}$ where $d_p$ and $\ell_p$ represent the dimension and length of the prompts respectively. Figure~\ref{fig2} illustrates the MMGM. Given $\boldsymbol{x}=(x^{am},x^v,x^t)$, we can generate the representation of the missing modality $x^{am}$ using the available $x^v$ and $x^t$ according to the following equation:
\begin{equation}\label{e1}
\hat x^a = f_{vt\rightarrow \hat a}([P_{Ga}, f_{v\rightarrow a}(x^v), f_{t\rightarrow a}(x^t)])
\end{equation}
where $\hat x^a$ denotes the representation generated, $[\dots]$ represents the concatenation operation, $f(\cdot)$ represents a Conv block which consists of a Conv 1D layer and an activation function and $\rightarrow$ represents from one or two modalities to another modality. If there are two missing modalities, such as $\boldsymbol{x}=(x^{am},x^{vm},x^t)$, the generation process is as follows:
\begin{equation}
\begin{aligned}
  \hat x^a = f_{t\rightarrow \hat a}([P_{Ga}, f_{t\rightarrow a}(x^t)]) \\
  \hat x^v = f_{t\rightarrow \hat v}([P_{Gv}, f_{t\rightarrow v}(x^t)]) 
\end{aligned}
\end{equation}
After applying the MMGM, we can represent the generated features as $\boldsymbol{x}=(\hat x^a,\hat x^v, x^t)$.

\begin{figure}
  \centering
  \includegraphics[width=0.9\linewidth]{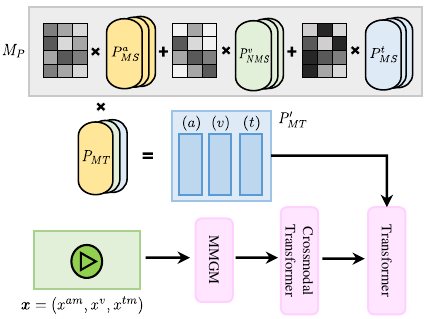}
  \caption{The illustration of attaching missing-type prompts to the Transformer. With the missing-type matrix $\mathbf{M_P}$, we generate missing-type prompts $P^\prime_{MT}$ for different missing modality cases. The figure shows the process of attaching missing-type prompts using an example of $\boldsymbol{x}=(x^{am},x^v,x^{tm})$ where audio and text modalities are missing.}
  \label{missing-type}
\end{figure}

\subsection{Missing-signal and Missing-type Prompts}\label{s33}
MMGM recovers missing information using available modalities. However, the information generated sometimes might not be accurate and could mislead the model. Therefore, missing-signal prompts are designed to inform the corresponding Transformer whether the information for a particular modality is real or generated. For each modality, there are two missing-signal prompts: $P_{MS}$ to denote a modality is missing and $P_{NMS}$ to denote a modality is not missing. As depicted in Figure~\ref{fig1}, after the MMGM and the Conv 1D layer, we obtain features $\boldsymbol{x}=(\hat x^a,x^v,x^t)$ where the audio modality is missing originally. We can incorporate the missing-signal prompts as follows:
\begin{equation}
  \begin{aligned}
    \hat x^a &:= \hat x^a + P_{MS}^a\\
    x^v &:= x^v + P_{NMS}^v\\
    x^t &:= x^t + P_{NMS}^t
  \end{aligned}
\end{equation}

After applying missing-signal prompts, the model knows which modalities are generated and which modalities are real, which can help the model make better use of the recovered information. Notably, missing-signal prompts are modality-specific which means that this kind of prompt only considers a specific modality and does not take into account the correlations between the absence of multiple modalities. To address this limitation, we propose missing-type prompts.

If there are $M$ modalities,  there can be a total of $2^M-1$ different cases of missing modalities. One intuitive approach is to design $2^M-1$ prompts to handle each situation individually~\citep{lee2023multimodal}. However, as the number of modalities increases, this approach becomes computationally expensive. Therefore, we introduce a missing-type projection matrix $\mathbf{M_P}$. We can obtain $\mathbf{M_P}$ of $\boldsymbol{x}=(x^{am},x^{v},x^{tm})$ as follows:
\begin{equation}
  \mathbf{M_P}=\mathbf{M_a}\cdot P_{MS}^a+\mathbf{M_v}\cdot P_{NMS}^v+\mathbf{M_t}\cdot P_{MS}^t
\end{equation}
where $\cdot$ is the matrix multiplication, $\mathbf{M_a}$, $\mathbf{M_v}$, $\mathbf{M_t}$$\in \mathbb R^{d_p\times \ell_p}$ and $\mathbf{M_p}\in \mathbb R^{d_p\times d_p}$. Then, we can get the missing-type prompts $P^\prime_{MT}$ as follows:
\begin{equation}
  P_{MT}^\prime = P_{MT}\cdot\mathbf{M_P}
\end{equation}
where $P_{MT}$ represents the original missing-type prompts, $P^\prime_{MT}$ represents the projected missing-type prompts and $P_{MT},P^\prime_{MT}\in\mathbb R^{3\times\ell_p\times d_p}$. In Figure~\ref{missing-type}, we illustrate how to attach the missing-type prompts to the Transformer with an example of $\boldsymbol{x}=(x^{am},x^{v},x^{tm})$. For each data pair in $\mathcal D$, the corresponding missing-type modality prompt is attached according to the situation of missing modalities.

\section{Experiments}
\subsection{Datasets and Evaluation Metrics}
To simulate real-world scenarios, we select CMU-MOSEI~\citep{bagher-zadeh-etal-2018-multimodal} as the high-resource dataset while CMU-MOSI~\citep{7742221}, IEMOCAP~\citep{Busso2008IEMOCAPIE} and CH-SIMS~\citep{yu-etal-2020-ch} are selected as the low-resource datasets. We pre-train our backbone on CMU-MOSEI and evaluate our proposed method on the four datasets.

\noindent\textbf{CMU-MOSI} is a popular dataset for multimodal (audio, text and video) sentiment analysis, comprising 93 English YouTube. Each segment is manually annotated with a sentiment score ranging from strongly negative to strongly positive (-3 to +3). 

\noindent\textbf{CMU-MOSEI} is an extension of CMU-MOSI. It contains more than 65 hours of annotated video from more than 1000 speakers and 250 topics. Compared with CMU-MOSI, it covers a wider range of topics.

\noindent\textbf{IEMOCAP} contains recorded videos from ten actors in five dyadic conversation sessions. It contains approximately 12 hours of data. Following previous works~\citep{wang2018words, tsai-etal-2019-multimodal}, four emotions (happiness, anger, sadness and neutral state) are selected for emotion recognition.

\noindent\textbf{CH-SIMS} is a Chinese multimodal sentiment analysis dataset. It contains 2,281 video segments annotated with a sentiment score ranging from strongly negative to strongly positive (-1 to 1). 

For CMU-MOSI and CMU-MOSEI, we follow previous works and adopt 7-class accuracy (ACC-7), binary accuracy (ACC), F1 score (F1), mean absolute error (MAE) and Pearson correlation (Corr) as evaluation metrics. For IEMOCAP, we implement four binary classification tasks and use the average accuracy (ACC) and F1-weighted score (F1) as evaluation metrics. For CH-SIMS, we use binary accuracy (ACC), F1 score (F1), mean absolute error (MAE) and Pearson correlation (Corr).

\begin{table*}[h!]
  \centering
  \caption{Quantitative results under six possible missing modality cases. For example, "$\{a\}$" means audio modality is available while video and text are missing. "Avg." means the average performance of the six possible cases. $\dagger$ denotes results copied from \citet{zhao-etal-2021-missing} where F1 score is not reported. \textbf{Bold}: best result. \underline{Underline}: second best result. We report the average result of five different random seeds.}
  \resizebox{\textwidth}{!}{%
  \begin{tabular}{@{}c|c|cccccccccccccc@{}}
  \toprule
  \multirow{2}{*}{Dataset} & \multirow{2}{*}{Method} & \multicolumn{2}{c}{$\{a\}$} & \multicolumn{2}{c}{$\{v\}$} & \multicolumn{2}{c}{$\{t\}$} & \multicolumn{2}{c}{$\{a,v\}$} & \multicolumn{2}{c}{$\{a,t\}$} & \multicolumn{2}{c}{$\{v,t\}$} & \multicolumn{2}{c}{Avg.} \\ \cmidrule(l){3-4} \cmidrule(l){5-6}\cmidrule(l){7-8}\cmidrule(l){9-10}\cmidrule(l){11-12}\cmidrule(l){13-14}\cmidrule(l){15-16}
  &  & ACC & F1 & ACC & F1 & ACC & F1 & ACC & F1 & ACC & F1 & ACC & F1 & ACC & F1 \\ \midrule
 \multirow{7}{*}{MOSI} & LB & 48.32 & 55.81 & 49.09 & 55.20 & 79.27 & 79.22 & 50.07 & 57.12 & 78.67 & 79.25 & 79.86 & 79.96 & 64.21 & 67.76 \\
  & MS & 49.17 & 55.34 & 49.87 & 56.12 & 78.06 & 78.28 & 51.12 & 57.01 & 79.32 & 79.65 & 80.32 & 80.38 & 64.64 & 67.80 \\
  & MD & 48.79 & 55.74 & 49.66 & 55.60 & 79.36 & 80.01 & 52.33 & 56.84 & 79.59 & 79.86 & 80.51 & 80.43 & 65.04 & 68.08 \\
  & MCTN & 51.32 & 56.12 & 54.27 & 56.33 & 79.63 & 79.78 & 56.79 & 57.84 & 78.96 & 79.17 & 80.45 & 80.65 & 66.90 & 68.32 \\
  & MMIN & {\ul 59.16} & {\ul 60.12} & {\ul 61.01} & {\ul 61.98} & {\ul 80.10} & {\ul 80.16} & {\ul 63.79} & {\ul 64.08} & {\ul 80.50} & {\ul 80.33} & 80.46 & 80.63 & {\ul 70.84} & {\ul 71.22} \\
  & MPMM & 57.26 & 59.35 & 58.63 & 59.12 & 79.81 & 80.10 & 60.54 & 61.33 & 79.89 & 79.84 & {\ul 80.74} & {\ul 80.93} & 69.48 & 70.11 \\
  & Ours & \textbf{62.71} & \textbf{63.65} & \textbf{63.12} & \textbf{63.74} & \textbf{80.12} & \textbf{80.31} & \textbf{65.02} & \textbf{65.41} & \textbf{80.76} & \textbf{81.09} & \textbf{81.12} & \textbf{81.19} & \textbf{72.14} & \textbf{72.57} \\ \midrule\midrule
 \multirow{7}{*}{IEMOCAP} & LB & 46.35 & 46.21 & 48.07 & 47.58 & 56.06 & 55.28 & 58.12 & 57.89 & 72.18 & 72.25 & 65.63 & 65.28 & 57.74 & 57.42 \\
  & MS & 47.65 & 47.52 & 47.68 & 47.36 & 59.27 & 59.22 & 57.48 & 56.60 & 72.30 & 72.18 & 66.81 & 66.93 & 58.53 & 58.30 \\
  & MD & 48.22 & 48.09 & 48.26 & 47.98 & 61.26 & 61.28 & 58.08 & 57.96 & 72.40 & 72.31 & 67.08 & 68.22 & 59.22 & 59.31 \\
  & MCTN & 51.62$\dagger$ & - & 45.73$\dagger$ & - & 63.78$\dagger$ & - & 55.84$\dagger$ & - & 69.46$\dagger$ & - & 68.34$\dagger$ & - & 59.19$\dagger$ & - \\
  & MMIN & {\ul 59.00}$\dagger$ & - & 51.60$\dagger$ & - & 68.02$\dagger$ & - & {\ul 65.43}$\dagger$ & - & {\ul 75.14}$\dagger$ & - & 73.61$\dagger$ & - & 65.47$\dagger$ & - \\
  & MPMM & 58.69 & 57.66 & {\ul 55.18} & {\ul 55.36} & {\ul 68.39} & {\ul 68.08} & 63.68 & 63.47 & 74.90 & 74.98 & 73.80 & 72.67 & {\ul 65.77} & {\ul 65.37} \\
  & Ours & \textbf{59.77} & \textbf{59.71} & \textbf{57.61} & \textbf{56.98} & \textbf{69.23} & \textbf{69.28} & \textbf{67.26} & \textbf{67.37} & \textbf{75.98} & \textbf{75.44} & \textbf{74.68} & \textbf{74.51} & \textbf{67.42} & \textbf{67.22} \\ \midrule\midrule
 \multirow{7}{*}{CH-SIMS} & LB & 63.82 & 75.15 & 64.08 & {\ul 78.11} & 76.74 & 76.90 & 62.14 & 73.21 & 76.84 & 76.93 & 77.01 & 77.13 & 70.11 & 76.24 \\
  & MS & 62.45 & 74.59 & 63.58 & 76.86 & 77.28 & 77.84 & 60.18 & 71.09 & 76.01 & 76.30 & 77.13 & 77.20 & 69.44 & 75.65 \\
  & MD & 64.22 & \textbf{77.25} & 63.87 & 76.01 & 77.34 & 77.48 & 62.91 & 72.14 & 76.77 & 76.92 & 77.14 & 77.31 & 70.38 & 76.19 \\
  & MCTN & 64.39 & 76.48 & 64.12 & 76.34 & 77.78 & 77.92 & 63.47 & 73.11 & 76.68 & 76.71 & 77.21 & 77.36 & 70.61 & 76.32 \\
  & MMIN & {\ul 65.21} & 77.09 & 65.32 & 77.41 & {\ul 78.91} & {\ul 78.67} & {\ul 64.28} & 73.36 & {\ul 77.32} & {\ul 77.33} & 77.40 & {\ul 77.48} & {\ul 71.41} & {\ul 76.89} \\
  & MPMM & 64.98 & 76.41 & {\ul 65.40} & 77.92 & 78.56 & 78.65 & 64.01 & {\ul 73.47} & 77.11 & 77.20 & {\ul 77.51} & 77.47 & 71.26 & 76.85 \\
  & Ours & \textbf{65.93} & {\ul 77.10} & \textbf{66.02} & \textbf{78.86} & \textbf{79.75} & \textbf{78.74} & \textbf{65.28} & \textbf{74.02} & \textbf{77.45} & \textbf{77.84} & \textbf{77.97} & \textbf{77.95} & \textbf{72.07} & \textbf{77.42} \\ \midrule\midrule
  \multirow{7}{*}{MOSEI} & LB & 66.21 & 68.69 & 66.45 & 69.10 & 77.96 & 78.32 & 67.30 & 69.62 & 78.13 & 78.63 & 77.86 & 78.16 & 72.32 & 73.83 \\
  & MS & 62.74 & 67.06 & 64.16 & 68.17 & 77.28 & 77.76 & 67.11 & 69.51 & 78.34 & 78.80 & 78.08 & 78.62 & 71.29 & 73.36 \\
  & MD & 65.76 & 68.18 & 66.57 & 69.35 & 77.30 & 77.94 & 67.21 & 69.48 & 78.74 & 78.97 & 78.11 & 78.71 & 72.28 & 73.82 \\
  & MCTN & 66.19 & 68.58 & 66.70 & 69.01 & 78.32 & 78.41 & 68.10 & 69.34 & 79.11 & 79.14 & 78.65 & 78.64 & 72.85 & 73.94 \\
  & MMIN & {\ul 67.11} & 68.67 & 67.01 & {\ul 69.31} & {\ul 78.67} & {\ul 78.71} & {\ul 68.17} & 69.74 & {\ul 79.94} & {\ul 79.96} & 79.32 & 79.29 & {\ul 73.37} & {\ul 74.39} \\
  & MPMM & 66.94 & \textbf{68.74} & {\ul 67.21} & 69.27 & 78.21 & 78.30 & 68.11 & {\ul 69.79} & 79.41 & 79.47 & {\ul 79.63} & {\ul 79.71} & 73.25 & 74.17 \\
  & Ours & \textbf{67.33} & {\ul 68.71} & \textbf{67.29} & \textbf{69.40} & \textbf{79.12} & \textbf{79.17} & \textbf{68.21} & \textbf{69.91} & \textbf{80.45} & \textbf{80.43} & \textbf{80.11} & \textbf{80.13} & \textbf{73.75} & \textbf{74.68} \\ \bottomrule
 \end{tabular}%
  }
  \label{table1}
  \vskip -0.1 in
\end{table*}

\subsection{Baselines}
We compare our proposed method with the following methods:
\textbf{Lower Bound (LB)} is trained with different combinations of modalities. Specifically, we train six different models using different combinations of modalities.
\textbf{Modality Substitution (MS)} substitutes missing modality with a default value or a placeholder.
\textbf{Modality Dropout (MD)} is a model trained with randomly dropped modalities during the training phase. 
\textbf{MCTN}~\citep{pham2020translation} learns robust joint representations by translating between modalities to deal with missing information.
\textbf{MMIN}~\citep{zhao-etal-2021-missing} learns robust joint multimodal representations, which can predict the representation of any missing modality given available modalities.
\textbf{MPMM}~\citep{lee2023multimodal} uses missing-aware prompts to instruct the model to address missing modality issues. 

\subsection{Implementation Details}
\textbf{Raw Feature Extraction.} To demonstrate the generalization ability of our method, we implement three kinds of methods to extract features. For CMU-MOSEI and CMU-MOSI, we follow \citet{tsai-etal-2019-multimodal} to extract features. For IEMOCAP, we follow \citet{zhao-etal-2021-missing} to extract acoustic, visual and textual features. For CH-SIMS, we follow \citet{yu-etal-2020-ch} to extract features.

\noindent\textbf{Model Training Details.} We first pre-train our backbone MulT~\citep{tsai-etal-2019-multimodal} on the CMU-MOSEI dataset. Then, we freeze all the parameters of the backbone and only train several learnable prompts, Conv layers and the output layer (As shown in Figure~\ref{fig1}). The length of prompts $\ell_p$ is set to 16 by default. We use L1 loss function for CMU-MOSEI, CMU-MOSI and CH-SIMS datasets and cross-entropy loss for IEMOCAP dataset. In all experiments, we use Adam optimizer with a batch size of 64. We train the model for 30 epochs with a learning rate of $1\times 10^{-3}$. Besides, we randomly discard the modality of the data with a missing rate of $\eta=70\%$ during training. We fix the random seed to ensure that each model is trained on the same data. 

\subsection{Main Results}
In Table~\ref{table1}, we present quantitative results on four datasets. The baselines LB, MS, MD and MPMM share the backbone of our method, thus reflecting the effectiveness of our proposed method to deal with missing modalities. Comparing the baseline MS and MD, we find that random discarding of data modalities during training improves the generalization ability of the model, thus making the model less sensitive to the data with missing modalities during the test phase.

Analyzing the results presented in the table, we observe that our proposed method outperforms the baselines by a large margin in all datasets under all six missing modality cases. Additionally, our method brings great enhancement when text modality is missing, with 8-13\% increase in accuracy compared with the LB baseline. This indicates that the three types of proposed prompts effectively guide the pre-trained model and yield impressive performance improvements. 

Besides, it is worth noting that we implement different feature extraction approaches on four datasets. From the results in Table~\ref{table1}, we can see that our model outperforms the baselines on all datasets, which shows that our model can adapt to features extracted by different methods. This indicates that our model learns relative rather than absolute relationships between features, demonstrating its robustness and versatility.

\begin{figure*}
  \centering
  \includegraphics[width=0.9\linewidth]{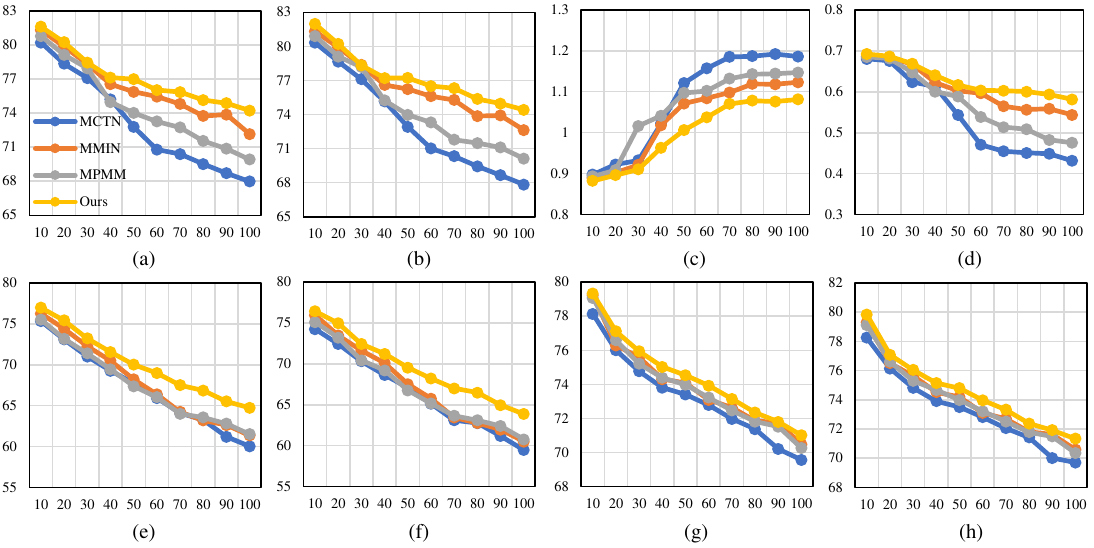}
  \caption{Performance comparison with different modality missing rates during tests. (a): ACC on CMU-MOSI. (b): F1 score on CMU-MOSI (c): MAE on CMU-MOSI. (d): Corr on CMU-MOSI. (e): ACC on IEMOCAP. (f): F1 score on IEMOCAP. (g): ACC on CH-SIMS. (h): F1 score on CH-SIMS.}
  \label{fig3}
  \vskip -0.1 in
\end{figure*}

In Figure~\ref{fig3}, we further compare the performance of our model with other methods under different modality missing rates during test time. From the figure, we find that our model performs better than all the other methods across all metrics under different modality missing rates, although the model is trained using the dataset with a modality missing rate of $\eta=70\%$. This indicates that our proposed method is robust to the missing rate of test set and can deal with severely missing modalities well.

Furthermore, the number of trainable parameters of our method is about 5-10\% percent of that of the backbone. The majority of trainable parameters come from the Conv layers in MMGM. The number of trainable parameters of three types of prompts only accounts for 0.5-1\% of that of the backbone. Notably, the number of trainable parameters does not increase with the size of the backbone network, which means that even if we use a much larger backbone, the number of trainable parameters remains the same. In all our experiments, we use only one 10 GB GPU (RTX 3080) with a batch size of 64. This demonstrates that our method is parameter-efficient.

\subsection{Generalization Ability}
To further validate the generalization ability of our method, we conduct experiments using different MSA/MER backbones. Specifically, we conduct experiments using MISA~\citep{hazarika2020misa}, MMIM~\citep{han-etal-2021-improving} and UniMSE~\citep{hu-etal-2022-unimse} and present the results in Table~\ref{back}. For all three backbones, we insert our generative prompts and module after the feature extractors. For UniMSE, we insert missing-signal and missing-type prompts into its multimodal fusion layers. For MMIM and MISA, we insert missing-signal and missing-type prompts into their modality-specific encoders and fusion layers, respectively. The results in the table demonstrate our method can enhance the ability of various backbones to address missing modality issues. Besides, our prompts can enhance the performance in the complete data situation, indicating that our missing-signal and missing-type prompts can help the model learn intra-modality and inter-modality features.

\begin{table}[]
  \centering
  \caption{Performance on different backbones on the CMU-MOSI dataset. "Com" denotes the complete data. "Incom" denotes the incomplete data. $\dagger$ denotes the method attached with prompts. For incomplete data, we report the average accuracy of six different missing conditions.}
  \label{back}
  \resizebox{0.95\columnwidth}{!}{%
  \begin{tabular}{@{}lcccc@{}}
  \toprule
  Backbone & Com & Com$\dagger$ & Incom & Incom$\dagger$ \\ \midrule
  MISA & 80.8 & 81.4$_{(+0.6)}$ & 67.9 & 73.4$_{(+5.5)}$  \\
  MMIM & 84.1 & 84.9$_{(+0.8)}$  & 68.6 & 72.3$_{(+3.7)}$  \\
  UniMSE & 86.9 & 87.4$_{(+0.5)}$  & 69.8 & 75.1$_{(+5.3)}$  \\ \bottomrule
  \end{tabular}%
  }
  \vskip -0.1 in
\end{table}

\subsection{Ablation Study}
We divide our ablation experiments into three parts: contributions of three types of prompts, the effect of modality missing rate during training and the effect of prompt length.

\begin{table*}[h!]
  \centering
  \caption{An ablation study on the benefit of the proposed generative prompts $\boldsymbol{P_G}$, missing-signal prompts $\boldsymbol{P_{MS}}$ and missing-type prompts $\boldsymbol{P_{MT}}$. $\checkmark$ represents a model with such type of prompts. \textbf{Bold}: best results with two kinds of prompts attached. * denotes best results with only one kind of prompt attached. We report the average performance of the six possible missing modality cases.}
  \resizebox{0.97\textwidth}{!}{%
  \begin{tabular}{@{}cccccccccccccc@{}}
  \toprule
  \multirow{2}{*}{$\boldsymbol{P_G}$} & \multirow{2}{*}{$\boldsymbol{P_{MS}}$} & \multirow{2}{*}{$\boldsymbol{P_{MT}}$} & \multicolumn{5}{c}{\textbf{CMU-MOSI}} & \multicolumn{4}{c}{\textbf{CH-SIMS}} & \multicolumn{2}{c}{\textbf{IEMOCAP}} \\ \cmidrule(l){4-8} \cmidrule(l){9-12}\cmidrule(l){13-14}
   &  &  & ACC-7 & ACC & F1 & MAE & Corr & ACC & F1 & MAE & Corr & ACC & F1 \\ \midrule
   &  &  & 27.41 & 65.04 & 68.08 & 1.121 & 0.549 & 70.38 & 76.19 & 0.555 & 0.506 & 59.22 & 59.31 \\
   $\checkmark$ &  &  & 28.27 & 70.26* & 71.33* & 1.097* & 0.574 & 71.08* & 76.57* & 0.518* & 0.521 & 64.12* & 63.97* \\
   & $\checkmark$ &  & 28.11 & 67.62 & 69.40 & 1.104 & 0.580* & 70.61 & 76.21 & 0.531 & 0.527* & 62.27 & 62.31 \\
   &  & $\checkmark$ & 30.17* & 66.41 & 69.21 & 1.167 & 0.561 & 70.49 & 76.13 & 0.542 & 0.515 & 63.48 & 63.20 \\ \midrule
   $\checkmark$& $\checkmark$ &  & 30.63 & \textbf{71.94} & \textbf{72.10} & 1.084 & 0.576 & \textbf{71.84} & \textbf{77.13} & \textbf{0.496} & 0.536 & 65.23 & 65.01 \\
   $\checkmark$&  & $\checkmark$ & 31.27 & 71.60 & 71.86 & \textbf{1.075} & 0.583 & 71.36 & 76.95 & 0.498 & 0.533 & \textbf{66.11} & \textbf{66.04} \\
   & $\checkmark$ & $\checkmark$ & \textbf{32.88} & 67.56 & 69.31 & 1.091 & \textbf{0.593} & 70.92 & 76.34 & 0.521 & \textbf{0.541} & 64.92 & 64.67 \\ \bottomrule
  \end{tabular}%
  }
  \label{table2}
  \vskip -0.1 in
\end{table*}

\begin{figure*}[h!]
  \centering
  \includegraphics[width=0.97\textwidth]{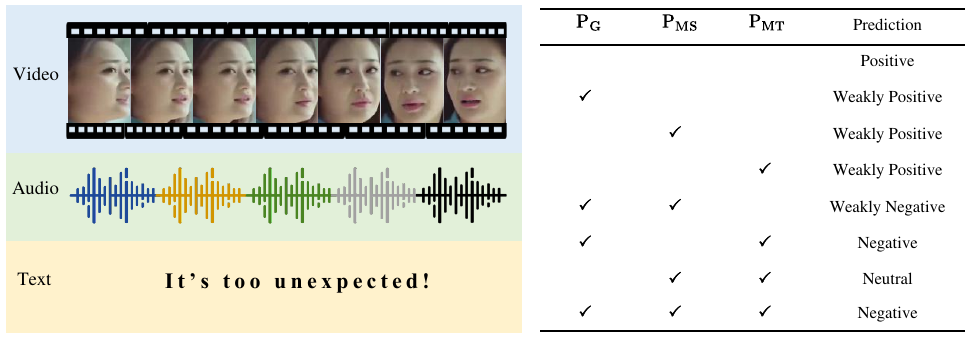}
  \caption{The effectiveness of three types of prompts on an example of CH-SIMS. The ground truth of the sample is "Negative". We report the results when the visual modality is missing.}
  \label{case}
  \vskip -0.1 in
\end{figure*}

\noindent\textbf{Contributions of three types of prompts.} In Table~\ref{table2}, we present quantitative results of the contributions of different prompts. For CMU-MOSI, we can observe that generative prompts give the greatest improvement in ACC and F1, while missing-signal prompts improve the Corr the most and missing-type prompts improve the ACC-7 the most. This indicates that generative prompts can help the available modalities generate the missing information which improves the binary accuracy. Besides, missing-type prompts tell the model whether other modalities are missing, thus strengthening the interactions between different modalities and learning cross-modal and fine-grained information which helps improve the ACC-7 a lot.

From the performance of models with different combinations of three types of prompts, we can further demonstrate the different roles of three types of prompts. We can conclude that generative prompts learn good representations of the missing modalities and improve the binary accuracy, missing-signal prompts are modality-specific prompts that tell models whether the corresponding modality is missing and help improve the correlation of the model's predictions with humans, and missing-type prompts are shared prompts with inter-modality information, thus helping models learn cross-modal and fine-grained information that improves ACC-7. Furthermore, the combinations of three types of prompts further enhance the performance of the model on all datasets. This fully confirms the validity of our proposed method. Besides, we use an example in CH-SIMS to study the effectiveness of three types of prompts and present the results in Figure~\ref{case}. From the figure, we can observe that the visual modality is key to predicting the correct result. With the prompts attached, the model can predict the accurate result, indicating the effectiveness of our method.

\begin{figure*}
  \centering
  \includegraphics[width=\textwidth]{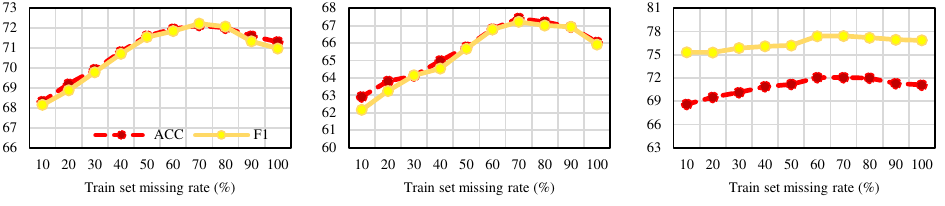}
  \caption{Quantitative results on CMU-MOSI (left), IEMOCAP (middle) and CH-SIMS (right) with different modality missing rates during training. We report the average performance under six different missing cases.}
  \label{fig4}
  \vskip -0.1 in
\end{figure*}

\begin{figure}[h!]
  \centering
  \includegraphics[width=0.92\linewidth]{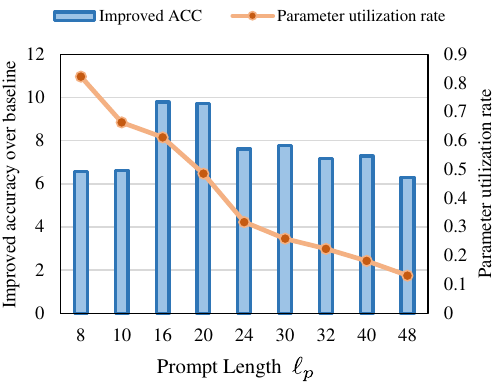}
  \caption{Quantitative results on CMU-MOSI with different prompt lengths $\ell_p$. The figure shows the improved accuracy (IACC) over the baseline Modality Dropout and the parameter utilization rate $\xi=\text{IACC}/\ell_p$.}
  \label{fig5}
  \vskip -0.15 in
\end{figure}

\noindent\textbf{The effect of modality missing rate during training.} We study the impact of modality missing rate during training on the performance of the model in Figure~\ref{fig4}. From the figure, we find that starting at a low point, both ACC and F1 score steadily improve as the train set modality missing rate increases, before reaching the highest point when the missing rate $\eta=70\%$. Then both ACC and F1 score decrease as the missing rate increases. This indicates that when the train set missing rate is low, it is difficult for a model to learn very good representations in the MMGM and to learn opportune prompts that can instruct the model well. This is because when the missing rate is low, the model tends to find a shortcut which to some degree prevents the model from learning good representations. With the missing rate higher, the model has to learn how to generate missing information to make predictions more accurate. However, if the missing rate is higher than 70\%, due to the amount of missing information, it is also hard for a model to learn good representations and prompts. 

\noindent\textbf{The effect of prompt length.} To study the impact of prompt length on our model, we train our model on CMU-MOSI with nine different prompt lengths and present results in Figure~\ref{fig5}. In the figure, we show the improved accuracy (IACC) over the baseline "Modality Dropout" of models with different prompt lengths. Intuitively, the longer the prompt length, the better the performance of the model. However, with the results shown in the figure, we find that when the prompt length $\ell_p=16$, the model performs the best. When the prompts are longer than 20, with the increase of the prompt length, the performance of the model decreases. Therefore, we deduce that it may be because our task is not complex and therefore the increase in parameters may overfit the model. Besides, we introduce parameter utilization rate $\xi=\text{IACC}/\ell_p$ to represent a trade-off between the performance of models and the number of parameters of prompts. From the figure, we can clearly see that $\ell_p=16$ is the best choice, where IACC and $\xi$ are both high compared with others. This also indicates that our proposed method can help improve the baseline with only a few parameters.

\section{Conclusion}
In this paper, we propose a novel multimodal Transformer via prompt learning to tackle the issue of missing modalities. We propose three types of prompts: generative prompts, missing-signal prompts, and missing-type prompts. Generative prompts can help generate missing information. Missing-signal prompts are modality-specific and missing-type prompts are modality-shared, which help the model learn intra-modality and inter-modality relationships respectively. 
With prompt learning, we can significantly reduce the number of trainable parameters. Extensive experiments and ablation studies demonstrate the effectiveness and robustness of our proposed method.

\section*{Limitations}
Our missing modality generation module generates missing information through two simple Conv blocks and generative prompts. This module improves the performance of our model significantly. However, we use extracted features but not raw features. Due to the simplicity of our missing generation module, the performance of the model could degrade if we use raw features which are much more complicated and have weaker correlation between modalities than extracted features. We leave this problem to future work.

\section*{Acknowledgements}
This work was supported by National Key R\&D Program of China under Grant No.2022ZD0162000.

\bibliography{main}

\end{document}